\documentclass{bmvc2k}

\title{Less is More: Facial Landmarks can Recognize a Spontaneous Smile}

\newcommand\blfootnote[1]{%
  \begingroup
  \renewcommand\thefootnote{}\footnote{#1}%
  \addtocounter{footnote}{-1}%
  \endgroup
}

\addauthor{Md. Tahrim Faroque Tushar$^\ast$}{tahrim.tushar@northsouth.edu}{1}
\addauthor{Yan Yang$^\ast$}{u6169130@anu.edu.au}{2,3}
\addauthor{Md Zakir Hossain}{zakir.hossain@anu.edu.au}{2,3,4}
\addauthor{Sheikh Motahar Naim}{sheinaim@amazon.com}{5}
\addauthor{Nabeel Mohammed}{nabeel.mohammed@northsouth.edu}{1}
\addauthor{Shafin Rahman}{shafin.rahman@northsouth.edu}{1$^{\dagger}$}

\addinstitution{
Department of Electrical and\\Computer Engineering,\\North South University, Bangladesh
}
\addinstitution{
 Biological Data Science Institute, \\ The Australian National University,\\
 Canberra, Australia
}

\addinstitution{
CSIRO Agriculture \& Food,\\
Canberra, Australia
}

\addinstitution{
CSIRO Machine Learning \& Artificial Intelligence Future Science Platform,\\
Canberra, Australia
}
\addinstitution{
Amazon Web Services
}

\runninghead{Tushar, Yan, Zakir, Sheikh, Nabeel, Shafin}{Less is More}

\def\etal{\emph{et al}\bmvaOneDot}

\usepackage{amsfonts}
\usepackage{booktabs}
\usepackage{multirow}
\usepackage{multicol}
\usepackage{sidecap}
\PassOptionsToPackage{hyphens}{url}\usepackage{hyperref}
\newcommand{\name}{\textbf{MeshSmileNet}}
\begin{document}

\maketitle
\blfootnote{$^\ast$ Equal Contribution.}
\blfootnote{$^{\dagger}$ Corresponding author.}

\begin{abstract}
Smile veracity classification is a task of interpreting social interactions. Broadly, it distinguishes between spontaneous and posed smiles. Previous approaches used hand-engineered features from facial landmarks or considered raw smile videos in an end-to-end manner to perform smile classification tasks. Feature-based methods require intervention from human experts on feature engineering and heavy pre-processing steps. On the contrary, raw smile video inputs fed into end-to-end models bring more automation to the process with the cost of considering many redundant facial features (beyond landmark locations) that are mainly irrelevant to smile veracity classification. It remains unclear to establish discriminative features from landmarks in an end-to-end manner. We present a \textbf{MeshSmileNet}~framework, 
a transformer architecture, to address the above limitations. To eliminate redundant facial features, our landmarks input is extracted from Attention Mesh, a pre-trained landmark detector. Again, to discover discriminative features, we consider the \textit{relativity} and \textit{trajectory} of the landmarks. For the \textit{relativity}, we aggregate facial landmark that conceptually formats a curve at each frame to establish local spatial features. For the \textit{trajectory}, we estimate the movements of landmark composed features across time by self-attention mechanism, which captures pairwise dependency on the trajectory of the same landmark.
This idea allows us to achieve state-of-the-art performances on UVA-NEMO, BBC, MMI Facial Expression, and SPOS datasets.
\end{abstract}

\section{Introduction}
\label{sec:intro}

Facial expression can reveal the emotional state of a person. One of the most common facial expressions representing happiness is the smile. More specifically, a genuine and spontaneous smile can only represent happiness. However, when the smile becomes deliberate or posed, it may signify anxiety, embarrassment, politeness, fear, affirmation and sarcasm ~\cite{ambadar2009all}. Due to the social impact of the smile, many studies have been conducted to recognize a genuine/real/spontaneous from a posed/deliberate/fake smile \cite{schmidt2009comparison, dibekliouglu2015recognition, wu2014spontaneous, yang2020realsmilenet}. Such research aims to improve human-computer interaction by allowing computers/robots to perceive users' emotional states. It can be helpful in the fields of neuroscience by studying the influence of subtle movements of eye and lip muscles in human emotions, psychology by investigating the genuine feeling of patients with mental illness, autism, or verbally handicapped people, criminology by aiding polygraph test for police interrogation and so on. In this paper, we study the spontaneous and posed smile expression and present a \name~framework that can work end-to-end and fully-automatic manner.

\begin{figure}[!]
\centering
\includegraphics[width=\linewidth]{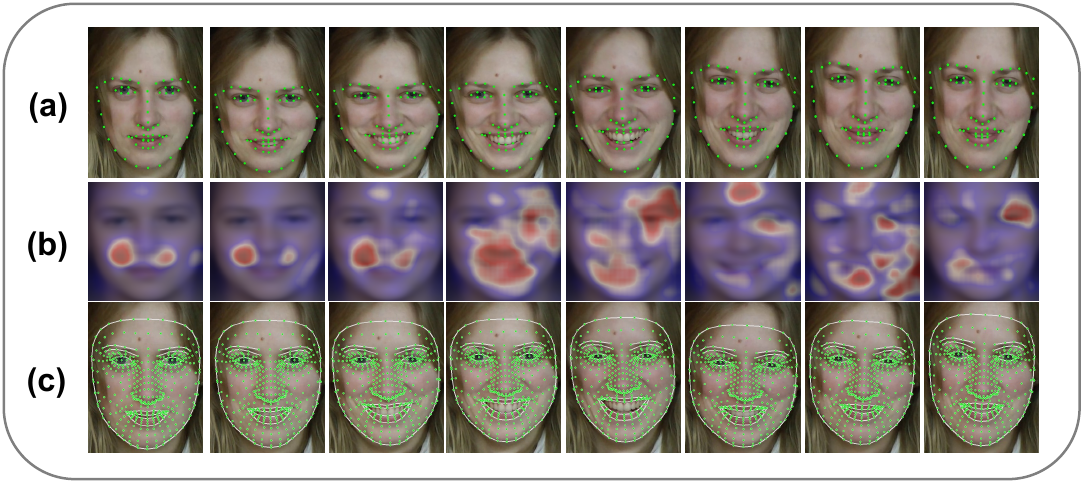}
\vspace{-2.5em}
\caption{\small Visual illustration of facial feature tracking of different methods related to smile classification. (a) \cite{pfister2011differentiating, valstar2006spontaneous, dibeklioglu2010eyes, dibekliouglu2012you, dibekliouglu2015recognition, wu2014spontaneous, mandal2017spontaneous} work on semi-automatic or automatic landmarks but extracts hand-engineered features. (b) \cite{mandal2016distinguishing, yang2020realsmilenet} use raw frames directly as input without any landmark location. Although they extract automatic features but need to consider redundant information. (c) Our approach contracts 3D landmark mesh to track down feature spatial and time dimensions automatically. It neither uses hand-engineered features nor redundant information of faces.}

  \label{fig:tracking1}
\end{figure}

Unlike traditional facial expression classification tasks, smile veracity classification is more challenging because of the difficulty of capturing and interpreting micro-facial movements in spatial and time dimensions. We identify several limitations of past methods. 
\textbf{(a)} \textit{Redundant facial features.}  Existing end-to-end learning approaches \cite{accv16, yang2020realsmilenet} study the problem by considering video frames as inputs. As indicated by \cite{dibekliouglu2012you}, the smile veracity mostly depends on the variation of key facial landmarks. For example, the pull of the zygomatic major and orbicularis oculi muscle leads to a spontaneous smile \cite{ekman1990duchenne}. However, video frames contain a large volume of irrelevant facial features such as identity that is trivial to the recognition performance.
\textbf{(b)} \textit{Poor inductive bias.} The spatial location of the same feature may vary across time dimensions. For example, the lips corner may not be in the similar location across different frames. For some methods, \cite{accv16,yang2020realsmilenet}, it creates difficulty in tracking feature trajectories because they assume an inductive bias that discriminative features will be located in a similar position across smile videos.
\textbf{(c)} \textit{Manual feature engineering.} Some methods \cite{dibekliouglu2012you} manually craft features including eye angle, cheek pull speed, and so on after tracking the facial landmark of interest over the smile video. Others \cite{wu2014spontaneous,wu2017spontaneous} explore Local Binary Pattern features for facial regions of interest. This process brings the intervention of human experts with strong domain knowledge and limits the feature generality to other similar tasks.
\textbf{(d)} \textit{Smile phase segmentation.} Manual feature engineering-based approach \cite{dibekliouglu2012you, wu2014spontaneous,wu2017spontaneous} usually requires the segmentation of the smile phase (onset, apex, offset) to underneath the recognition performance. It exacerbates the cumbersome recognition processes of these feature engineering-based approaches.
\textbf{(e)} \textit{Spatial isolation.} Manually crafted features are processed by a single-layer model \cite{dibeklioglu2010eyes,dibekliouglu2012you,dibekliouglu2015recognition,wu2014spontaneous}, where the interactions of features from different facial regions are strictly treated as linear. This strong assumption hurts the recognition performance because the complex interplay of smile features should be predominantly non-linear. Figure \ref{fig:tracking1} summarizes facial feature tracking on different methods.

We present a \name~framework that avoids redundant facial features and inject inductive bias of facial movements by converting each frame of the smile video into facial landmarks. We extract landmarks using Attention Mesh \cite{grishchenko2020attention}, a pre-trained model, frame by frame. Landmarks as input (instead of raw frames) make the input representation compact and enable landmark tracking across the time dimension. To yield discriminative landmark features,  we explore the concept of
\textit{relativity} and \textit{trajectory} for the smile veracity classification task. 
For the \textit{relativity},  we process landmarks locally in the spatial domain with a sequence of CurveNet block \cite{xiang2021walk}. It selectively groups the nearby landmarks that interact with each other. Moreover, it conceptually approximates geometrical landmark features, for example, the eye angle from those manual feature-based approaches. 
The spatial relativity of each landmark provides efficient features variation tracking compared to raw landmarks because of containing the nearby landmark information.
For the \textit{trajectory}, movements for the same landmark feature across time are accounted by a temporal self-attention mechanism. We place a lightweight spatial self-attention to avoid spatial isolation. The spontaneous smile usually has slow facial feature movements than the posed smile \cite{schmidt2006movement}. In time domains, we ground feature variation by self-attention (e.g., the speed of pulling lip corners). 
With this idea, our \name~model extracts effective landmark features for the smile veracity classification. 
We perform extensive experiments and achieve state-of-the-art results on UVA-NEMO, BBC, MMI facial expression, and SPOS datasets.

Our contributions are listed below: 
\textit{\textbf{(1)}} We propose the \name~framework for classifying spontaneous and posed smiles based on facial landmarks. 
\textit{\textbf{(2)}} We explore the concepts, relativity and trajectory, of learning end-to-end landmark features for smile classification task.  
\textit{\textbf{(3)}} We perform detailed ablation studies, compare our methods with strong baselines and state-of-the-art results on four well-known smile datasets.


\section{Related Works}

\textbf{Smile Recognition.} In the facial action coding system (FACS)~\cite{ekman1978facial,bartlett2006automatic}, a smile triggers both the Orbicularis oculi (AU6) and the Zygomatic major (AU12) muscle by rising cheeks and lip corners. In regards to the smile duration,  Dibeklioglu \etal \cite{dibeklioglu2010eyes,dibekliouglu2015recognition} separate it into onset (neutral to expressive), apex (stay expressive), and offset (decay to neutral) phases. Later works on smile motions are studied to distinguish spontaneous and posed smiles. Valstar \etal \cite{valstar2006spontaneous,valstar2007distinguish} studies the eyebrows motion patterns and movements of head, face, and shoulders. Dibeklioglu \etal \cite{dibeklioglu2010eyes,dibekliouglu2015recognition} present D-Marker features that consider eyelids, cheeks, and lip corners movements. Mandal \etal \cite{mandal2017spontaneous} distinguish spontaneous smiles from posed smiles by dense optical flows.
Pfister \etal \cite{pfister2011differentiating} and Wu \etal \cite{wu2014spontaneous,wu2017spontaneous} present a spatial-temporal texture description, CLBP-TOP, to encode eye, lip, and cheek for identifying spontaneous smile. However, all the above methods are semi-automatic that requires human intervention for feature engineering. Mandal \etal \cite{mandal2016distinguishing}, and Yang \etal \cite{yang2020realsmilenet} study automatic spontaneous and posed smile recognition with convolution networks by smile videos. However, they consider redundant information from raw video frames that slowdowns the efficiency of the classification process. Our paper proposes a novel automatic framework that studies the relativity and trajectory of facial landmarks for spontaneous and posed smile recognition to address the issues of past methods.

\noindent\textbf{Relating smiles with video and point cloud classification.} 
Recent advancements in video and point cloud sequence classification approaches can benefit real/fake smile recognition methods. 
Existing transformer-based approaches for video classification focus on the study of patch tokenization, positional embedding, feedforward network activation, normalization, and efficient self-attention to scale to high volume video classification \cite{srv}. A standard approach uses a convolution network to extract spatial features of each frame and aggregate them by self-attention in the time domain. In our proposed method, we adopt such self-attention. Furthermore, to classify smile videos, one can convert a smile video to facial landmarks with 4D points and apply point cloud classification methods. Fan \etal~\cite{fan2022pstnet,fan2021deep} propose PSTNet and P4Transformer with a spatial and temporal convolution to learn informative feature representation of the point cloud sequences. By a fixed spatial radius and time span, this convolution aggregates features in nearby spatial locations and time separately. In PSTNet, they progressively downsample point features in the spatial domain with the furthest point sample to perceive the entire point cloud. P4Transformer combines PSTNet with transformers to have a global interaction in the time domain for features at close spatial locations. The spatial feature aggregations of both PSTNet and P4Transformer consider all nearby landmarks. In our task, nearby landmarks are not equally important. They potentially flatten beneficial features for spontaneous and posed smile recognition. Our framework explores the relativity of nearby landmarks to help the task.

\section{Method}

\subsection{Problem Formulation}
From psychological research, we know that lips, cheeks, and eyes play important roles in a genuine smile \cite{dibeklioglu2010eyes,dibekliouglu2012you,dibekliouglu2015recognition}. For this, traditional methods find landmark locations in semi-automatic or automatic ways and track the movement of those locations using some pre-defined rules/equations/statistics (e.g., piecewise Bézier volume deformation tracker and speed, acceleration, duration, symmetries and amplitude of landmarks \cite{dibeklioglu2010eyes,dibekliouglu2012you,dibekliouglu2015recognition}).
In some cases, methods \cite{wu2014spontaneous,wu2017spontaneous,pfister2011differentiating} extract low-level vision features (e.g., HoG, LBP)  surrounding landmark locations. We consider traditional approaches as hand-engineered feature-based methods because it requires expert intervention. Recent deep learning approaches \cite{mandal2016distinguishing, yang2020realsmilenet} make feature extraction automatic (replacing hand-engineered features) by using whole raw video frames as input. Since such approaches use CNN features without considering landmark information, they need to consider redundant facial information unrelated to smiles. Moreover, the spatial location of key landmarks might not be similar across diverse human subjects (e.g., young vs. adults), which may restrict the network from extracting necessary information. Our goal is to propose a deep architecture that can track landmark locations in an end-to-end manner without using redundant information and expert hand-engineering. For this, we leverage 3D landmarks obtained from Attention Mesh~\cite{grishchenko2020attention}.

Given a video $\mathcal{X} = [\mathbf{x}_n \mid n \in 1 \ldots N]$, where $\mathbf{x}_t$ and $N$ are $t$th frame and video length, respectively, we aim to identify if $\mathcal{X}$ is a spontaneous $(y = 0)$ or posed  $(y = 1)$ smile. Here, $y$ is the true label. From each frame, $\mathbf{x}_{n} \in \mathcal{X}$, we extract 3D facial landmarks, $\mathbf{a}_{n} \in \mathbb{R}^{3 \times L}$, where $L$ is the total number of landmarks. Suppose, $\mathcal{A} = [\mathbf{a}_{n} \mid n \in 1 \ldots N]$ denotes the landmark representation of a smile video. Our goal is to train a model $\mathcal{F}(\cdot)$ that maps $\mathcal{A}$ to $y$. 

\subsection{\name}

\begin{figure}[!] \label{Fig:2}
\centering
\includegraphics[width=\linewidth,trim={0cm 0cm 0cm 0cm},clip]{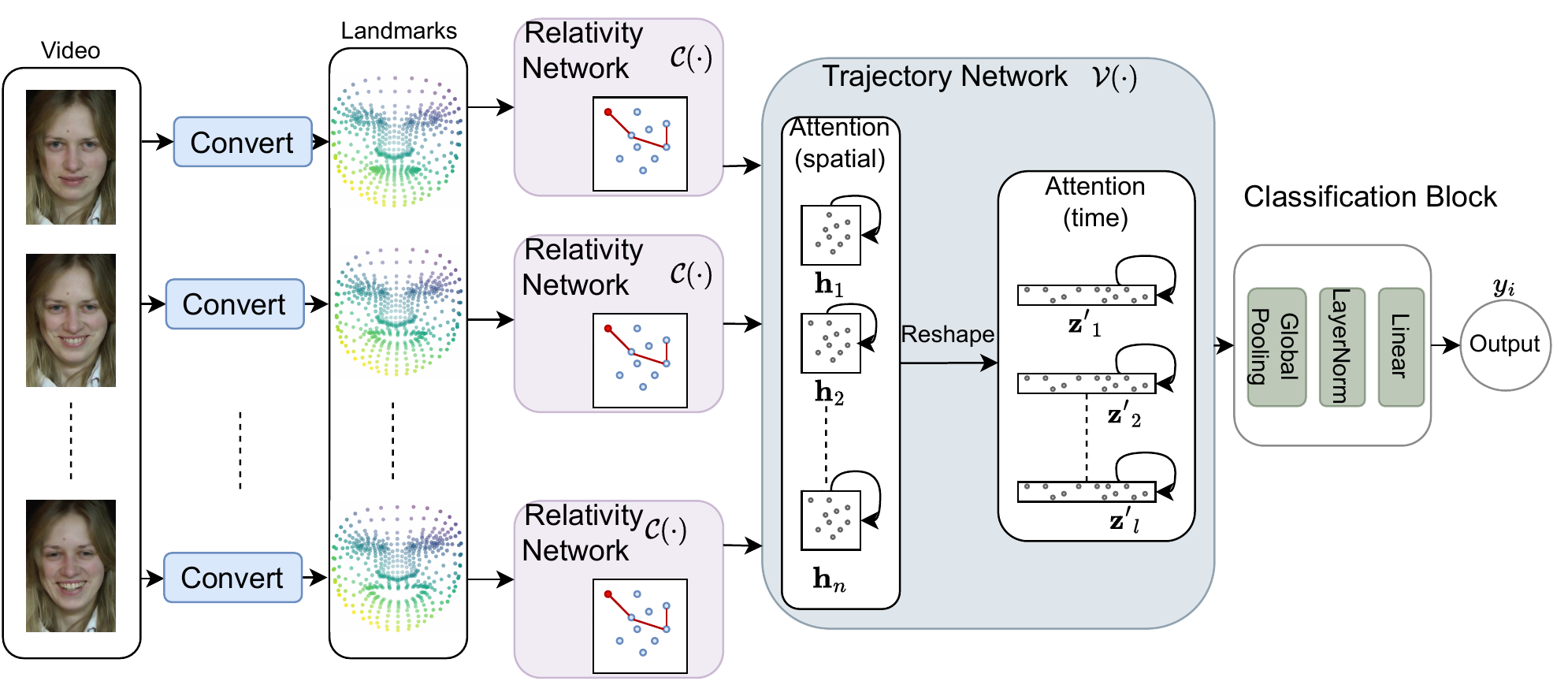}
\vspace{-2em}
  \caption{\small Our proposed architecture, \name. See supplementary material for more details.
  }
  \label{fig:architecture}
\end{figure}

\noindent\textbf{Model Overview.}
First, we extract $L$ number of facial landmarks $(\mathbf{a}_{n})$ from each frame $(\mathbf{x}_n)$ of a video using Attention Mesh~\cite{grishchenko2020attention}. As shown in Figure \ref{fig:architecture}, all extracted landmarks are fed into our proposed network, $\mathcal{F}(\cdot)$.
It composes of three main components.
\textbf{(1)} Relativity network: We explore spatial geometry relations of landmarks  at each frame ($\mathbf{a}_{n} \in \mathcal{A}$) based on CurveNet blocks \cite{xiang2021walk}. It links landmarks with similar features into a curve before aggregation.
\textbf{(2)} Trajectory network: As $\mathbf{a}_{n} \in \mathcal{A}$ is ordered, we track the movement of each landmark $\mathbf{a}_{n,l}$, 
$l$th landmark at $n$th frame. This operation is based on the self-attention mechanism, aiming to measure the variation of the constructed geometry feature across time. To have interactions between the learned geometric features (avoid  spatial isolation), we employ a lightweight self-attention for them on spatial domain. It also helps trajectory measurement of our purpose. 
\textbf{(3)} Classification network: We classify the video into label $y$, spontaneous smiles or posed smiles, from the relativity and trajectory based landmark features. Now, we elaborate on each network's components.

\noindent\textbf{Relativity network.}  Our relativity network $\mathcal{C}(\cdot)$ models the local geometry of nearby landmarks. Past studies show geometry-based features such as lip pulling, which is formed by a curve from at least three landmarks, are beneficial in the spontaneous and posed smile recognition task. We borrow CurveNet blocks \cite{xiang2021walk} into our framework for this purpose. For landmarks of the same frame,  a CurveNet block forms $k$ curves. It determines their starting points by top-k selection method \cite{gao2019graph} that ranks projected scores of each landmark. Afterward, a curve is sequentially formulated by connecting to neighborhood landmarks with Gumbel-Softmax liked selection methods before pooling into a unified size of landmark features.
We denote these processes by:
\begin{equation*}
    \mathbf{h}_{n} = \mathcal{C}(\mathbf{a}_{n}) \quad \forall n \in [1\cdots N],
\end{equation*}
where, $h_{n} \in \mathbb{R}^{d \times L}$ is $d$-dimensional output representation of L landmarks.  


\noindent\textbf{Trajectory network.} We measure the variation of the learned geometry feature by the trajectory network $\mathcal{V}(\cdot)$. It builds up representations that describe the movements of facial features such as the pull of the zygomatic major and orbicularis oculi muscle. Naively, it could be achieved by introducing a self-attention to all landmarks. However, this is computationally expensive. Motivated by the axial attention \cite{axt}, we decompose it into self-attention in spatial and time domains. Our $\mathcal{V}(\cdot)$ has 
\begin{align*}
    &&\mathbf{o}_{n} &= \text{Attention}(\mathbf{h}_{n},\mathbf{h}_{n},\mathbf{h}_{n}) \quad&& \forall n \in [1\cdots N], \\
    &&\mathbf{z'}_{l} &= [\mathbf{o}_{1,l} \cdots \mathbf{o}_{N,l}] \quad&& \forall l \in [1\cdots L], \\
    &&\mathbf{z}_{l} &= \text{Attention}(\mathbf{z'}_{l},\mathbf{z'}_{l},\mathbf{z'}_{l}) \quad&& \forall l \in [1\cdots L],   
\end{align*}
where, $\mathbf{o}_{n} \in \mathbb{R}^{d \times L}$ is the output for attention on each frame, $\mathbf{z'}_{l} \in \mathbb{R}^{d \times N}$ concatenates feature of $l$th landmark across time, and $\mathbf{z}_{l} \in \mathbb{R}^{d \times N}$ is the attention output for, $\mathbf{z}_{l}$, the same landmark across time. $\mathbf{o}_{n}$ facilitates better spatial local feature representation with interacting spatial landmark features, and $\mathbf{z}_{l}$ tracks the movement of local facial features. Semantically, following on the above example, $\forall l \in [1\cdots L]~\mathbf{z}_{l}$ can model `the pull of the zygomatic major muscle' or `the pull of the orbicularis oculi muscle' separately. Here, $\forall n \in [1\cdots N]~\mathbf{o}_{n}$ brings interactions by aggregating them. 

\begin{SCfigure}
    \centering
    \includegraphics[width=.55\linewidth]{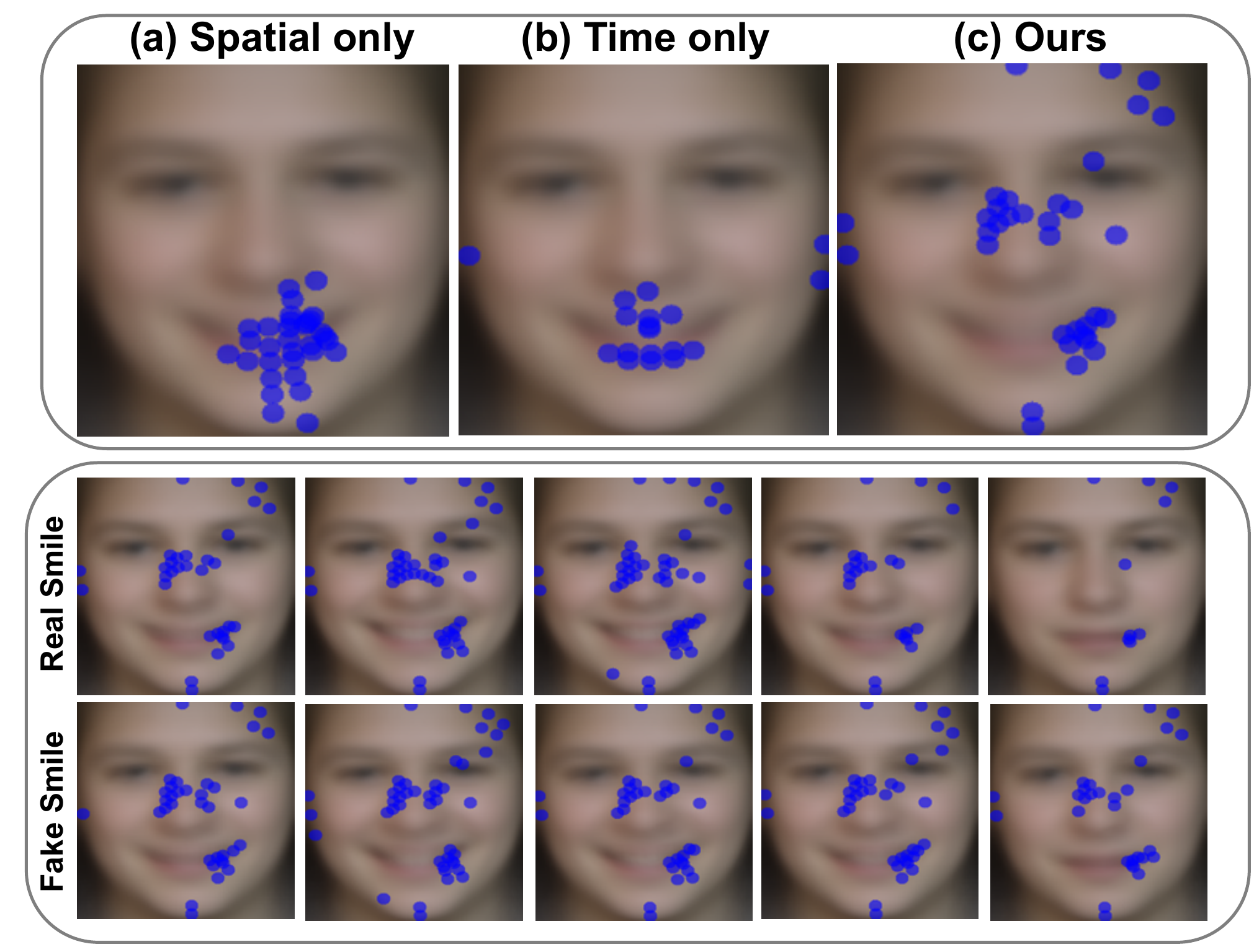}
    \caption{\small Overall distinguishing face landmarks (marked as blue) for real fake smile classification while (a) only spatial, (b) only time dimensional, and (c) both spatial and time dimensional self-attention is used on UVA-NEMO dataset. Our method finds landmarks related to lips, cheeks, eyes, and face borders essential for the classification task. The bottom two rows show distinguishing face landmarks of five individual samples of real and fake smile videos, respectively. It tells that important landmarks are found in similar locations regardless of real or fake smiles.
    }
    \label{fig:3}
\end{SCfigure}

In Figure \ref{fig:3}, we analyze the complex interplay of relativity and trajectory networks by visualizing the importance of different landmark locations by calculating their gradient with respect to the target label. Note that we average the background face, landmark position, and gradient of the entire UVA-NEMO dataset. The trajectory network, $\mathcal{V}(\cdot)$, calculates self-attention across spatial and time dimensions of videos. While using only spatial self-attention in Figure \ref{fig:3}(a), we notice that lips regions (spanning from nose bottom to chin) play decisive facial features for the task. Then, only self-attention across time dimensions in Figure \ref{fig:3}(b) also finds the landmarks around the left and right sides of the face border. By using self-attention across both spatial and time dimensions, our final recommendation in Figure \ref{fig:3}(c) combines individual contributions (of spatial and time) and additionally focuses on the nose and eyes regions. This finding is somewhat similar to Dibeklioglu \textit{et al.} \cite{dibekliouglu2015recognition} which establishes that lips, cheek, and eyes contribute more to real and fake smile classification. Our method automatically arrives at the similar conclusion without any hand-engineering. Being data dependent nature of deep learning, our method also concentrates on the nose and face border-related landmarks.

\noindent\textbf{Classification Block.} We apply a global pooling on $\forall l \in [1\cdots L]~\mathbf{z}_{l}$, and then it is passed to distinguish between spontaneous and posed smiles. Our Classification Block has a global pooling layer, a Layer Normalization, a Linear layer, and a Sigmoid activation function.

\noindent\textbf{Objective:}  
Our goal is to classify spontaneous and posed smiles. We penalize deviation of network predictions $\mathcal{F}(\mathcal{A})$ from the ground truth $y_{i}$ using Binary Cross Entropy loss:
\begin{equation*} \label{eq:3}
\mathcal{L}_{BCE}= y_i\log\mathcal{F}(\mathcal{A}) + (1-y_i)\log(1-\mathcal{F}(\mathcal{A}))
\end{equation*}

\noindent\textbf{Inference.} Given a test smile video, after extracting landmarks locations representing that video, $\mathcal{A^*}$, we perform a forward pass for prediction, $\hat{y}$, i.e., $\hat{y} = \mathcal{F}(\mathcal{A^*})$.

\section{Experiments}
\textbf{Datasets.} We experiment with four smile datasets: 
\textbf{(a)} UvA-NEMO dataset\cite{dibekliouglu2015recognition}: This dataset contains 597 spontaneous and 643 posed smile videos of 400 participants. It is recorded in resolution $1920 \times 1080$ at 50 frames per second. Among the four datasets, this one has a higher resolution and frames per second (FPS) and the highest number of videos. 
\textbf{(b)} BBC dataset: This dataset has 20 videos, divided into 10 posed and 10 spontaneous. The videos are recorded in resolution $314 \times 286$ with 25 FPS. 
\textbf{(c)} MMI facial expression dataset\cite{valstar2010induced}: It contains spontaneous and posed facial expressions. Similar to \cite{dibekliouglu2012you,yang2020realsmilenet}, we use 
138 spontaneous and 49 posed smile videos. The spontaneous smile part is recorded in resolution $640 \times 480$ and at 29 FPS, and the posed smile videos are recorded in $720 \times 576$ pixels at 25 FPS. 
\textbf{(d)} SPOS dataset \cite{pfister2011differentiating}: There is two types of image sequences, gray and near-infrared sequences and all the images are in $640 \times 480$ pixels with 25 FPS. There are in total 80 smiles, 66 spontaneous, and 14 deliberate smiles. We use the gray image sequences in our research. Detailed statistics of smile datasets are presented in the supplementary material.

\noindent\textbf{Train/test split \& evaluation.}  We follow the train/test split and 10-cross-validation settings from \cite{dibekliouglu2015recognition} for the UvA-NEMO dataset. For the BBC, MMI, and SPOS datasets, we follow \cite{dibekliouglu2015recognition, wu2014spontaneous} and use 10-fold, 9-fold, and 7-fold cross-validation, respectively. All the datasets maintain no subject overlap in training and testing splits. To evaluate the model, we use prediction accuracy. We run ten trials to report the average result.

\noindent\textbf{Implementation details\footnote{Codes are available at: \sloppy{https://github.com/Tushar-Faroque/MeshSmileNet}}.}
We extract 478 landmarks per frame with Attention Mesh~\cite{grishchenko2020attention}. Our model is trained with these extracted landmarks. We use a batch size of 16 and train the model for 300 epochs. We use AdamW optimizer with a learning rate of $5e-4$. We sample 10 FPS for the UvA-NEMO, MMI, and BBC dataset and 1 FPS for the SPOS dataset. In training, we sample 16 continuous frames per video. During testing, we average the prediction score for 5 samples per video. In the Relativity Network, we use a CurveNet block having four CIC layers \cite{xiang2021walk}. We have a convolution 1d layer to linearly mix the 478 landmark features per frame to 32 features and then pass it to the Trajectory network $\mathcal{V}(\cdot)$. It has nine regular transformer blocks. Six of them perform spatial self-attention, and three are for self-attention in the time domain. We use the \textit{PyTorch} library to perform our experiments.

\begin{table}[!t]
\begin{minipage}{0.5\textwidth}
\resizebox{\textwidth}{!}{
\begin{tabular}{ccccc}
\toprule
Method & UVA-NEMO & MMI & SPOS & BBC \\
\midrule
Cohn'04~\cite{cohn2004timing} & 77.3 & 81.0 & 73.0 & 75.0 \\
Dibeklioglu'10~\cite{dibeklioglu2010eyes} & 71.1 & 74.0 & 68.0 & 85.0 \\
Pfister'11~\cite{pfister2011differentiating} & 73.1 & 81.0 & 67.5 & 70.0 \\
Wu'14~\cite{wu2014spontaneous} & 91.4 & 86.1 & 79.5 & 90.0 \\
Dibeklioglu'15~\cite{dibekliouglu2015recognition} & 89.8 & 88.1 & 77.5 & 90.0 \\
Wu'17~\cite{wu2017spontaneous} & \textbf{93.9} & 92.2 & 81.2 & 90.0 \\
Mandal'17~\cite{mandal2017spontaneous} & 80.4 & - & - & - \\
\hline
Mandal'16~\cite{mandal2016distinguishing} & 78.1 & - & - & - \\
RealSmileNet'20~\cite{yang2020realsmilenet} & 82.1 & 92.0 & 86.2 & 90.0 \\
\hline
PSTNet~\cite{fan2022pstnet} & 72.9 & 94.3 & 87.1 & 95.0 \\
P4Transformer~\cite{fan2021point} & 74.9 & 91.3 & 82.9 & 85.0 \\
Vanilla ViT \cite{vit} & 78.4 & 99.0 & 93.5 & 95.0 \\
\hline
Ours & 85.0 & \textbf{99.0} & \textbf{94.4} & \textbf{95.0}\\
\bottomrule
\end{tabular}}
\caption{\small Comparison of state-of-the-art methods. 
}
\label{tab:1}
\end{minipage}
\begin{minipage}{0.48\textwidth}


\begin{center}
\resizebox{\textwidth}{!}
{\begin{tabular}{ccccc}
\toprule
Method & UVA-NEMO & MMI & SPOS & BBC \\
\midrule
Dibeklioglu'12~\cite{dibekliouglu2012you} & 72.1 & 78.2 & 53.2 & 55.0 \\
Dibeklioglu'15~\cite{dibekliouglu2015recognition} & 77.0 & 73.9 & 46.9 & 55.0 \\
\hline
Ours & \textbf{85.0 }& \textbf{99.0} & \textbf{94.4} & \textbf{95.0}\\
\bottomrule
\end{tabular}}
\end{center}
\vspace{-0.8em}
\caption{\centering\small{Using Attention Mesh landmark extractors on other related methods \cite{dibekliouglu2012you, dibekliouglu2015recognition}.
\label{tab:landmarkcomp}
}}
\centering
\scalebox{.73}{
\resizebox{\textwidth}{!}{
\begin{tabular}{ccc}
\toprule
Individual Group & Real & \\
UVA-NEMO & SmileNet~\cite{yang2020realsmilenet} & Ours \\
\midrule
Young & 79.6 & \textbf{80.4} \\
Adult & 79.4 & \textbf{82.4} \\
Male &  77.8 & \textbf{81.4} \\
Female & 80.0 & \textbf{80.2} \\
\bottomrule
\end{tabular}}}

\caption{\centering\small{Results on diverse subject groups.}}
\label{tab:subjectgroup}
\end{minipage}
\end{table}

\subsection{Main Results}

\textbf{Comparison methods.} We compare our work with three categories of methods. 
(a) Semi-automatic: These approaches \cite{pfister2011differentiating, valstar2006spontaneous, dibeklioglu2010eyes, dibekliouglu2012you, dibekliouglu2015recognition, wu2014spontaneous, mandal2017spontaneous} require manual annotation of facial lankmarks for the first frame in a smile video. Also, the extract smile features are hand-engineered by experts.
(b) Fully-automatic: Unlike the semi-automatic case, these approaches \cite{mandal2016distinguishing, yang2020realsmilenet} require no manual annotation for the first frame. They use CNN layers on the raw video frames which allows many redundant smile features.
(c) Baseline: We propose three baselines that could be a reasonable but incomplete version of our method. After extracting landmarks from each frame using Attention Mesh \cite{grishchenko2020attention}, we represent a video by 4D point clouds. Then, we apply point cloud sequence classification methods like PSTNet~\cite{fan2022pstnet} and P4Transformer~\cite{fan2021point} to classify smile videos. Also, we feed raw video frames to the vanilla vision transformer (ViT) \cite{vit} model for video classification. PSTNet, P4Transformer, and ViT represent three baselines of our method.

\noindent\textbf{Performance benchmark.} In Table \ref{tab:1}, we present a benchmark comparison of our methods with state-of-the-art approaches based on four datasets. We separate
`Semi-automatic', `Fully-Automatic', and `Baseline' methods by horizontal lines. Our observations are as follows:
\textbf{(1)} Compared to semi-automatic methods, we achieve state-of-the-art results in the MMI, SPOS, and BBC datasets. In the UvA-NEMO dataset, the best performance was reported by Wu \etal~\cite{wu2017spontaneous}, but it requires hand-crafted features and heavy preprocessing steps. Our model is the most competitive automatic method in classifying spontaneous smiles, which classifies spontaneous smiles from facial landmarks with a single forward pass. 
\textbf{(2)} When comparing past fully automatic methods, our method significantly outperforms them. Mandal \etal~\cite{mandal2016distinguishing} and Yang \etal~\cite{yang2020realsmilenet} classify spontaneous smile with a CNN that inputs raw frames. However, in this task, most facial features are irrelevant, such as identity information, but they redundantly cover key facial landmarks that are beneficial for the spontaneous smile classification task. As a result, their approaches are difficult to understand the spontaneous smile-related features. Unlike them, our \name~resolves the difficulty by learning relativity and trajectory of landmarks extracted from raw frames.
\textbf{(3)} Our model outperforms baseline methods by a huge margin. Again, the vanilla ViT applies self-attention to raw frames in the spatial domain and time domain. Though global interactions are achieved, variations of the same spatial features need to be modeled from many patch features. It creates difficulty for the spontaneous smile classification. PSTNet~\cite{fan2022pstnet} and P4Transformer~\cite{fan2021point} have landmark inputs. They use spatial and temporal convolution and P4Transformer~\cite{fan2021point} to aggregate features from nearby landmarks. In our task, landmarks are not equally important to each other. Our \name~selects nearby landmarks interacting with each other before aggregation and improves spontaneous smile classification performance.

\begin{table}[!t]
\begin{minipage}{0.6\textwidth}
\begin{center}
\resizebox{\textwidth}{!}{
\begin{tabular}{ccccccc}
\hline
Input & CurveNet & Attention & UVA-NEMO & MMI & SPOS & BBC \\
\hline
Frames & No & Both$^*$ & 78.4 & 99.0 & 93.5 & 95.0 \\
Landmarks & No & Both$^*$ & 82.2 & 96.7 & 92.2 & 95.0 \\
Landmarks & Yes & Spatial & 72.3 & 98.5 & 90.1 & 95.0 \\
Landmarks & Yes & Time & 82.4 & 98.5 & 90.9 & 95.0 \\
Landmarks & Yes & Both$^*$ & \textbf{85.0} & \textbf{99.0} & \textbf{94.4} & \textbf{95.0} \\
\hline
\end{tabular}}
\end{center}
\vspace{-0.8em}
\caption{\centering\small Ablation on model architectures. $^*$ means self-attention is used in both spatial and time direction. 
}
\label{tab:u}
\end{minipage}
\begin{minipage}{0.38\textwidth}
\resizebox{\textwidth}{!}{
\begin{tabular}{ccccc}
\toprule
Frame Rate & UVA-NEMO & MMI & SPOS & BBC \\
\midrule
1 & 67.0 & 98.6 & \textbf{94.4} & \textbf{95.0} \\
3 & 74.5 & 98.6 & 90.3 & \textbf{95.0} \\
5 & \textbf{85.0} & \textbf{99.0} & 92.4 & \textbf{95.0} \\
10 & 82.2 & 97.6 & 92.4 & 90.0 \\
\bottomrule
\end{tabular}}
\vspace{-1.0em}
\caption{\centering\small Impact of frame rates acroess difference datasets.
}
\label{tab:f}
\end{minipage}
\end{table}

\noindent\textbf{Significant test.} We follow \cite{sigte} to determine if our methods significantly outperform the automatic baselines. With the 10-fold partitions from the UVA-NEMO (each testing fold contains around 100 videos, and there are 1240 testing videos in total), we use the paired t-test to compare our methods against the baselines in Table \ref{tab:sigte}. Since p-values are $<10^{-3}$, we conclude that our method exhibits statistically significant results than the baselines.

\noindent\textbf{Role of landmark extractor.} We work on FaceMash, which extracts more landmark locations (478 vs. 11) than the landmark tracker from past methods. In Table \ref{tab:landmarkcomp}, we compare the performance of using FaceMash landmarks on Dibeklioglu'12 \cite{dibekliouglu2012you} and  Dibeklioglu'15~\cite{dibekliouglu2015recognition}. Their feature extraction process is extremely sensitive to landmark locations because they usually track facial landmarks across the video with a pre-defined set of rules and statistics. Moreover, they require manual annotation of landmarks in the first frame. 
Here, when scaling their approach using the automatic landmark extractor, FaceMash, they perform lower than our proposed model. It tells that more landmarks do not help until features are learned end-to-end using deep learning methods (instead of hand-engineered features).

\noindent\textbf{Subject group result.} UVA-NEMO dataset provides smile data of different subject groups, e.g., young, adult, male and female. We apply our method in each individual groups and compare the result with \textit{RealSmileNet}~\cite{yang2020realsmilenet} in Table \ref{tab:subjectgroup}. Our method outperforms \cite{yang2020realsmilenet} which confirms the robustness of the proposed method. This becomes possible because of excluding redundancies of raw frames which were used in  \cite{yang2020realsmilenet}.

\subsection{Ablation studies}

\noindent \textbf{Ablation on architecture.} 
In Table \ref{tab:u}, we perform ablation studies based on architecture components. Firstly, we feed raw frames directly to the ViT model for video classification where no CurveNet is used . It achieves poor performance because of redundant features. Secondly, we directly apply landmark inputs to the vanilla ViT model with self-attention in both spatial and time domains. It addresses redundant feature issues and shows considerable performance improvement. Thirdly, we include the relativity network (implemented as CurveNet), which deals with spatial geometry relations of landmarks. Then, we separately try spatial and time self-attention in the trajectory network. For both cases, we notice performance improvement over the second case because of measuring the variation of the constructed geometry feature through the trajectory network. Finally, we apply all components as our final recommended method, outperforming all previously ablated architectures. Using landmarks as input instead of raw frames and the interplay between relativity and trajectory networks help our method achieve outstanding results.

\begin{table}[!t]
    \begin{minipage}{0.55\linewidth}
    \centering
    \resizebox{\linewidth}{!}{
    \begin{tabular}{cccc}
        \toprule
          & Ours vs. ViT &  Ours vs. PSTNet & Ours vs. P4Transformer \\
         \midrule
         t-statistic & 8.87 & 22.97& 10.38\\
         p-value & $9.6\times 10^{-6}$ &$7.5\times10^{-8}$& $2.6 \times 10^{-6}$\\
         \bottomrule
    \end{tabular}}
    \vspace{-1em}
    \caption{Significance tests with baselines.}
    \label{tab:sigte}
    \end{minipage}
    \hspace{2em}
    \begin{minipage}{0.35\linewidth}
    \centering
    \vspace{1em}
    \resizebox{\linewidth}{!}{
    \begin{tabular}{cccc}
        \toprule
         Method & PSTNet & P4Transformer & Ours  \\
         \midrule
         DLIB \cite{king2009dlib}  & 69.7 & 62.9 & 80.3\\
         Attention Mesh \cite{grishchenko2020attention} & \textbf{72.9} & \textbf{74.9} & \textbf{85.0}\\
         \bottomrule
    \end{tabular}}
    \vspace{-1em}
    \caption{DLIB vs. Attention Mesh landmark extractor.}
    \label{tab:dli}
    \end{minipage}
\end{table}

\noindent \textbf{Ablation on frame rate.} We study the impact of frame rates (Table \ref{tab:f}) for our model on four datasets. Using 5 FPS per video, our model performs the best in UVA-NEMO, MMI, and BBC. However, 1 FPS serves the best for the SPOS. Note SPOS dataset provides only the onset phase of the smile videos. In contrast, UVA-NEMO, MMI, and BBC provide all three phases (onset, apex, and offset) in the smile videos. The onset phase usually is in slow motion. Thus, low FPS finds the best performance for SPOS. Nevertheless, FPS to be used can be considered as a hyper-parameter for validation.

\noindent \textbf{Ablation on landmark extractor.} We compare Attention Mesh landmark extractor \cite{grishchenko2020attention} with widely used DLIB extractor \cite{king2009dlib} (with 68 landmark locations). In Table \ref{tab:dli}, we show the results on automatic baselines and with our method. We exclude ViT from the comparison since ViT does not use facial landmarks. Because of providing more landmark locations in Attention Mesh (478 vs. 68), it beats the DLIB extractor. However, our method performs well against the baselines for both extractors' cases, showing our generalization ability.

\section{Conclusion}
In this paper, we propose a framework for real/fake smile recognition, which leverages facial landmarks from video data to learn spatial and temporal. In the past, models relied on raw smile frame input, manual feature engineering, phase segmentation, etc., which hampers the efficiency of the overall smile classification. In contrast, we focus on the relatively and trajectory of facial landmarks using CurveNet and the self-attention mechanism network, respectively. This way, we learn spatial and temporal features from them and classify spontaneous and posed smiles. We executed experiments on UvA-NEMO, MMI, BBC, and SPOS smile datasets and achieve state-of-the-art results.

\noindent\textbf{Acknowledgement.} This work was supported by North South University (NSU) Conference Travel and Research Grants (CTRG) 2021–2022 (Grant ID: CTRG-21-SEPS-10).
{
\small
\bibliography{egbib}

\begin{thebibliography}{31}
\providecommand{\natexlab}[1]{#1}
\providecommand{\url}[1]{\texttt{#1}}
\expandafter\ifx\csname urlstyle\endcsname\relax
  \providecommand{\doi}[1]{doi: #1}\else
  \providecommand{\doi}{doi: \begingroup \urlstyle{rm}\Url}\fi

\bibitem[Ambadar et~al.(2009)Ambadar, Cohn, and Reed]{ambadar2009all}
Zara Ambadar, Jeffrey~F Cohn, and Lawrence~Ian Reed.
\newblock All smiles are not created equal: Morphology and timing of smiles
  perceived as amused, polite, and embarrassed/nervous.
\newblock \emph{Journal of nonverbal behavior}, 33\penalty0 (1):\penalty0
  17--34, 2009.

\bibitem[Bartlett et~al.(2006)Bartlett, Littlewort, Frank, Lainscsek, Fasel,
  Movellan, et~al.]{bartlett2006automatic}
Marian~Stewart Bartlett, Gwen Littlewort, Mark~G Frank, Claudia Lainscsek,
  Ian~R Fasel, Javier~R Movellan, et~al.
\newblock Automatic recognition of facial actions in spontaneous expressions.
\newblock \emph{J. Multim.}, 1\penalty0 (6):\penalty0 22--35, 2006.

\bibitem[Cohn and Schmidt(2004)]{cohn2004timing}
Jeffrey~F Cohn and Karen~L Schmidt.
\newblock The timing of facial motion in posed and spontaneous smiles.
\newblock \emph{International Journal of Wavelets, Multiresolution and
  Information Processing}, 2\penalty0 (02):\penalty0 121--132, 2004.

\bibitem[Dibeklioglu et~al.(2010)Dibeklioglu, Valenti, Salah, and
  Gevers]{dibeklioglu2010eyes}
Hamdi Dibeklioglu, Roberto Valenti, Albert~Ali Salah, and Theo Gevers.
\newblock Eyes do not lie: Spontaneous versus posed smiles.
\newblock In \emph{Proceedings of the 18th ACM international conference on
  Multimedia}, pages 703--706, 2010.

\bibitem[Dibeklio{\u{g}}lu et~al.(2012)Dibeklio{\u{g}}lu, Salah, and
  Gevers]{dibekliouglu2012you}
Hamdi Dibeklio{\u{g}}lu, Albert~Ali Salah, and Theo Gevers.
\newblock Are you really smiling at me? spontaneous versus posed enjoyment
  smiles.
\newblock In \emph{European Conference on Computer Vision}, pages 525--538.
  Springer, 2012.

\bibitem[Dibeklio{\u{g}}lu et~al.(2015)Dibeklio{\u{g}}lu, Salah, and
  Gevers]{dibekliouglu2015recognition}
Hamdi Dibeklio{\u{g}}lu, Albert~Ali Salah, and Theo Gevers.
\newblock Recognition of genuine smiles.
\newblock \emph{IEEE Transactions on Multimedia}, 17\penalty0 (3):\penalty0
  279--294, 2015.

\bibitem[Dietterich(1998)]{sigte}
Thomas Dietterich.
\newblock Approximate statistical tests for comparing supervised classification
  learning algorithms.
\newblock \emph{Neural computation}, 10:\penalty0 1895--1923, 10 1998.

\bibitem[Dosovitskiy et~al.(2020)Dosovitskiy, Beyer, Kolesnikov, Weissenborn,
  Zhai, Unterthiner, Dehghani, Minderer, Heigold, Gelly, Uszkoreit, and
  Houlsby]{vit}
Alexey Dosovitskiy, Lucas Beyer, Alexander Kolesnikov, Dirk Weissenborn,
  Xiaohua Zhai, Thomas Unterthiner, Mostafa Dehghani, Matthias Minderer, Georg
  Heigold, Sylvain Gelly, Jakob Uszkoreit, and Neil Houlsby.
\newblock An image is worth 16x16 words: Transformers for image recognition at
  scale.
\newblock 10 2020.

\bibitem[Ekman and Friesen(1978)]{ekman1978facial}
Paul Ekman and Wallace~V Friesen.
\newblock Facial action coding system.
\newblock \emph{Environmental Psychology \& Nonverbal Behavior}, 1978.

\bibitem[Ekman et~al.(1990)Ekman, Davidson, and Friesen]{ekman1990duchenne}
Paul Ekman, Richard~J Davidson, and Wallace~V Friesen.
\newblock The duchenne smile: Emotional expression and brain physiology: Ii.
\newblock \emph{Journal of personality and social psychology}, 58\penalty0
  (2):\penalty0 342, 1990.

\bibitem[Fan et~al.(2021{\natexlab{a}})Fan, Yang, and
  Kankanhalli]{fan2021point}
Hehe Fan, Yi~Yang, and Mohan Kankanhalli.
\newblock Point 4d transformer networks for spatio-temporal modeling in point
  cloud videos.
\newblock In \emph{Proceedings of the IEEE/CVF Conference on Computer Vision
  and Pattern Recognition}, pages 14204--14213, 2021{\natexlab{a}}.

\bibitem[Fan et~al.(2021{\natexlab{b}})Fan, Yu, Yang, and
  Kankanhalli]{fan2021deep}
Hehe Fan, Xin Yu, Yi~Yang, and Mohan Kankanhalli.
\newblock Deep hierarchical representation of point cloud videos via
  spatio-temporal decomposition.
\newblock \emph{IEEE Transactions on Pattern Analysis and Machine
  Intelligence}, 2021{\natexlab{b}}.

\bibitem[Fan et~al.(2022)Fan, Yu, Ding, Yang, and Kankanhalli]{fan2022pstnet}
Hehe Fan, Xin Yu, Yuhang Ding, Yi~Yang, and Mohan Kankanhalli.
\newblock Pstnet: Point spatio-temporal convolution on point cloud sequences.
\newblock \emph{arXiv preprint arXiv:2205.13713}, 2022.

\bibitem[Gao and Ji(2019)]{gao2019graph}
Hongyang Gao and Shuiwang Ji.
\newblock Graph u-nets.
\newblock In \emph{international conference on machine learning}, pages
  2083--2092. PMLR, 2019.

\bibitem[Grishchenko et~al.(2020)Grishchenko, Ablavatski, Kartynnik,
  Raveendran, and Grundmann]{grishchenko2020attention}
Ivan Grishchenko, Artsiom Ablavatski, Yury Kartynnik, Karthik Raveendran, and
  Matthias Grundmann.
\newblock Attention mesh: High-fidelity face mesh prediction in real-time.
\newblock \emph{arXiv preprint arXiv:2006.10962}, 2020.

\bibitem[Ho et~al.(2019)Ho, Kalchbrenner, Weißenborn, and Salimans]{axt}
Jonathan Ho, Nal Kalchbrenner, Dirk Weißenborn, and Tim Salimans.
\newblock Axial attention in multidimensional transformers.
\newblock 12 2019.

\bibitem[King(2009)]{king2009dlib}
Davis~E King.
\newblock Dlib-ml: A machine learning toolkit.
\newblock \emph{The Journal of Machine Learning Research}, 10:\penalty0
  1755--1758, 2009.

\bibitem[Mandal and Ouarti(2017)]{mandal2017spontaneous}
Bappaditya Mandal and Nizar Ouarti.
\newblock Spontaneous versus posed smiles—can we tell the difference?
\newblock In \emph{Proceedings of International Conference on Computer Vision
  and Image Processing}, pages 261--271. Springer, 2017.

\bibitem[Mandal et~al.(2016{\natexlab{a}})Mandal, Lee, and Ouarti]{accv16}
Bappaditya Mandal, David Lee, and Nizar Ouarti.
\newblock Distinguishing posed and spontaneous smiles by facial dynamics.
\newblock In Chu{-}Song Chen, Jiwen Lu, and Kai{-}Kuang Ma, editors,
  \emph{Computer Vision - {ACCV} 2016 Workshops - {ACCV} 2016 International
  Workshops, Taipei, Taiwan, November 20-24, 2016, Revised Selected Papers,
  Part {I}}, volume 10116 of \emph{Lecture Notes in Computer Science}, pages
  552--566. Springer, 2016{\natexlab{a}}.
\newblock \doi{10.1007/978-3-319-54407-6\_37}.
\newblock URL \url{https://doi.org/10.1007/978-3-319-54407-6\_37}.

\bibitem[Mandal et~al.(2016{\natexlab{b}})Mandal, Lee, and
  Ouarti]{mandal2016distinguishing}
Bappaditya Mandal, David Lee, and Nizar Ouarti.
\newblock Distinguishing posed and spontaneous smiles by facial dynamics.
\newblock In \emph{Asian Conference on Computer Vision}, pages 552--566.
  Springer, 2016{\natexlab{b}}.

\bibitem[Pfister et~al.(2011)Pfister, Li, Zhao, and
  Pietik{\"a}inen]{pfister2011differentiating}
Tomas Pfister, Xiaobai Li, Guoying Zhao, and Matti Pietik{\"a}inen.
\newblock Differentiating spontaneous from posed facial expressions within a
  generic facial expression recognition framework.
\newblock In \emph{2011 IEEE International Conference on Computer Vision
  Workshops (ICCV Workshops)}, pages 868--875. IEEE, 2011.

\bibitem[Schmidt et~al.(2006)Schmidt, Ambadar, Cohn, and
  Reed]{schmidt2006movement}
Karen~L Schmidt, Zara Ambadar, Jeffrey~F Cohn, and L~Ian Reed.
\newblock Movement differences between deliberate and spontaneous facial
  expressions: Zygomaticus major action in smiling.
\newblock \emph{Journal of nonverbal behavior}, 30\penalty0 (1):\penalty0
  37--52, 2006.

\bibitem[Schmidt et~al.(2009)Schmidt, Bhattacharya, and
  Denlinger]{schmidt2009comparison}
Karen~L Schmidt, Sharika Bhattacharya, and Rachel Denlinger.
\newblock Comparison of deliberate and spontaneous facial movement in smiles
  and eyebrow raises.
\newblock \emph{Journal of nonverbal behavior}, 33\penalty0 (1):\penalty0
  35--45, 2009.

\bibitem[Selva et~al.(2022)Selva, Johansen, Escalera, Nasrollahi, Moeslund, and
  Clapés]{srv}
Javier Selva, Anders Johansen, Sergio Escalera, Kamal Nasrollahi, Thomas
  Moeslund, and Albert Clapés.
\newblock Video transformers: A survey.
\newblock 01 2022.

\bibitem[Valstar et~al.(2010)Valstar, Pantic, et~al.]{valstar2010induced}
Michel Valstar, Maja Pantic, et~al.
\newblock Induced disgust, happiness and surprise: an addition to the mmi
  facial expression database.
\newblock In \emph{Proc. 3rd Intern. Workshop on EMOTION (satellite of LREC):
  Corpora for Research on Emotion and Affect}, page~65. Paris, France., 2010.

\bibitem[Valstar et~al.(2006)Valstar, Pantic, Ambadar, and
  Cohn]{valstar2006spontaneous}
Michel~F Valstar, Maja Pantic, Zara Ambadar, and Jeffrey~F Cohn.
\newblock Spontaneous vs. posed facial behavior: automatic analysis of brow
  actions.
\newblock In \emph{Proceedings of the 8th international conference on
  Multimodal interfaces}, pages 162--170, 2006.

\bibitem[Valstar et~al.(2007)Valstar, Gunes, and
  Pantic]{valstar2007distinguish}
Michel~F Valstar, Hatice Gunes, and Maja Pantic.
\newblock How to distinguish posed from spontaneous smiles using geometric
  features.
\newblock In \emph{Proceedings of the 9th international conference on
  Multimodal interfaces}, pages 38--45, 2007.

\bibitem[Wu et~al.(2017)Wu, Liu, Zhang, and Gao]{wu2017spontaneous}
Ping-ping Wu, Hong Liu, Xue-wu Zhang, and Yuan Gao.
\newblock Spontaneous versus posed smile recognition via region-specific
  texture descriptor and geometric facial dynamics.
\newblock \emph{Frontiers of Information Technology \& Electronic Engineering},
  18\penalty0 (7):\penalty0 955--967, 2017.

\bibitem[Wu et~al.(2014)Wu, Liu, and Zhang]{wu2014spontaneous}
Pingping Wu, Hong Liu, and Xuewu Zhang.
\newblock Spontaneous versus posed smile recognition using discriminative local
  spatial-temporal descriptors.
\newblock In \emph{2014 IEEE International Conference on Acoustics, Speech and
  Signal Processing (ICASSP)}, pages 1240--1244. IEEE, 2014.

\bibitem[Xiang et~al.(2021)Xiang, Zhang, Song, Yu, and Cai]{xiang2021walk}
Tiange Xiang, Chaoyi Zhang, Yang Song, Jianhui Yu, and Weidong Cai.
\newblock Walk in the cloud: Learning curves for point clouds shape analysis.
\newblock In \emph{Proceedings of the IEEE/CVF International Conference on
  Computer Vision}, pages 915--924, 2021.

\bibitem[Yang et~al.(2020)Yang, Hossain, Gedeon, and
  Rahman]{yang2020realsmilenet}
Yan Yang, Md~Zakir Hossain, Tom Gedeon, and Shafin Rahman.
\newblock Realsmilenet: a deep end-to-end network for spontaneous and posed
  smile recognition.
\newblock In \emph{Proceedings of the Asian Conference on Computer Vision},
  2020.

\end{thebibliography}
}
\end{document}